\documentclass[11pt,a4paper]{article}
\usepackage[hyperref]{acl2020}
\usepackage{times}
\usepackage{latexsym}

\usepackage{graphicx}
\usepackage{amsmath,amsfonts,amssymb}
\usepackage{multirow}
\usepackage{url}
\usepackage{footnote}

\usepackage{microtype}

\aclfinalcopy % Uncomment this line for the final submission
\title{Dense-Caption Matching and Frame-Selection Gating for \\ Temporal Localization in VideoQA}

\author{Hyounghun Kim \;\;\;\;\;\;\;
Zineng Tang \;\;\;\;\;\;\;
Mohit Bansal \\
UNC Chapel Hill \\
  {\tt \{hyounghk, terran, mbansal\}@cs.unc.edu}}

\date{}

\begin{document}
\maketitle
\begin{abstract}
Videos convey rich information. Dynamic spatio-temporal relationships between people/objects, and diverse multimodal events are present in a video clip. Hence, it is important to develop automated models that can accurately extract such information from videos. Answering questions on videos is one of the tasks which can evaluate such AI abilities. In this paper, we propose a video question answering model which effectively integrates multi-modal input sources and finds the temporally relevant information to answer questions. Specifically, we first employ dense image captions to help identify objects and their detailed salient regions and actions, and hence give the model useful extra information (in explicit textual format to allow easier matching) for answering questions. Moreover, our model is also comprised of dual-level attention (word/object and frame level), multi-head self/cross-integration for different sources (video and dense captions), and gates which pass more relevant information to the classifier. Finally, we also cast the frame selection problem as a multi-label classification task and introduce two loss functions, In-and-Out Frame Score Margin (IOFSM) and Balanced Binary Cross-Entropy (BBCE), to better supervise the model with human importance annotations. We evaluate our model on the challenging TVQA dataset, where each of our model components provides significant gains, and our overall model outperforms the state-of-the-art by a large margin (74.09\% versus 70.52\%). We also present several word, object, and frame level visualization studies.\footnote{Our code is publicly available at: \url{https://github.com/hyounghk/VideoQADenseCapFrameGate-ACL2020}}

\end{abstract}

\section{Introduction}
Recent years have witnessed a paradigm shift in the way we get our information, and a lot of it is related to watching and listening to videos that are shared in huge amounts via the internet and new high-speed networks. Videos convey a diverse breadth of rich information, such as dynamic spatio-temporal relationships between people/objects, as well as events. Hence, it has become important to develop  automated models that can accurately extract such precise multimodal information from videos~\cite{tapaswi2016movieqa,maharaj2017dataset,kim2017deepstory,jang2017tgif, gao2017tall, anne2017localizing, lei2018tvqa,lei2019tvqa+}.
Video question answering is a representative AI task through which we can evaluate such abilities of an AI agent to understand, retrieve, and return desired information from given video clips. 

In this paper, we propose a model that effectively integrates multimodal information and locates the relevant frames from diverse, complex video clips such as those from the video+dialogue TVQA dataset~\cite{lei2018tvqa}, which contains questions that need both the video and the subtitles to answer. When given a video clip and a natural language question based on the video, naturally, the first step is to compare the question with the content (objects and keywords) of the video frames and subtitles, then combine information from different video frames and subtitles to answer the question. Analogous to this process, we apply dual-level attention in which a question and video/subtitle are aligned in word/object level, and then the aligned features from video and subtitle respectively are aligned the second time at the frame-level to integrate information for answering the question. Among the aligned frames (which contain aggregated video and subtitle information now), only those which contain relevant information for answering the question are needed. Hence, we also apply gating mechanisms to each frame feature to select the most informative frames before feeding them to the classifier. 

Next, in order to make the frame selection more effective, we cast the frame selection sub-task as a multi-label classification task. To convert the time span annotation to the label for each frame, we assign a positive label (`1') to frames between the start and end points, and negative (`0') label to the others, then train them with the binary cross-entropy loss. Moreover, for enhanced supervision from the human importance annotation, we also introduce a new loss function, In-and-Out Frame Score Margin (IOFSM), which is the difference in average scores between in-frames (which are inside the time span) and out-frames (which are outside the time span). We empirically show that these two losses are complementary when they are used together. Also, we introduce a way of applying binary cross-entropy to the unbalanced dataset. As we see each frame as a training example (positive or negative), we have a more significant number of negative examples than positive ones. To balance the bias, we calculate normalized scores by averaging the loss separately for each label. This modification, which we call balanced binary cross-entropy (BBCE), helps adjust the imbalance and further improve the performance of our model.

Finally, we also employ dense captions to help further improve the temporal localization of our video-QA model. Captions have proven to be helpful for vision-language tasks \cite{wu2019generating, Li_2019_CVPR, kim-bansal-2019-improving} by providing additional, complementary information to the primary task in descriptive textual format. We employ dense captions as an extra input to our model since dense captions describe the diverse salient regions of an image in object-level detail, and hence they would give more useful clues for question answering than single, non-dense image captions. 

Empirically, our first basic model (with dual-level attention and frame-selection gates) outperforms the state-of-the-art models on TVQA validation dataset (72.53\% as compared to 71.13\% previous state-of-the-art) and with the additional supervision via the two new loss functions and the employment of dense captions, our model gives further improved results (73.34\% and 74.20\% respectively). These improvements from each of our model components (i.e., new loss functions, dense captions) are statistically significant. Overall, our full model's test-public score substantially outperforms the state-of-the-art score by a large margin of 3.57\% (74.09\% as compared to 70.52\%).\footnote{At the time of the ACL2020 submission deadline, the publicly visible rank-1 entry was 70.52\%. Since then, there are some new entries, with results up to 71.48\% (compared to our 74.09\%).} Also, our model's scores across all the 6 TV shows are more balanced than other models in the TVQA leaderboard\footnote{{\scriptsize{https://competitions.codalab.org/competitions/20415\#results}}}, implying that our model should be more consistent and robust over different genres/domains that might have different characteristics from each other. 

Our contributions are four-fold: (1) we present an effective model architecture for the video question answering task using dual-level attention and gates which fuse and select useful spatial-temporal information, (2) we employ dense captions as salient-region information and integrate it into a joint model to enhance the videoQA performance by locating proper information both spatially and temporally in rich textual semi-symbolic format, (3) we cast the frame selection sub-task as a multi-level classification task and introduce two new loss functions (IOFSM and BBCE) for enhanced supervision from human importance annotations (which could be also useful in other multi-label classification settings), and (4) our model's score on the test-public dataset is 74.09\%, which is around 3.6\% higher than the state-of-the-art result on the TVQA leaderboard (and our model's scores are more balanced/consistent across the diverse TV show genres).  We also present several ablation and visualization analyses of our model components (e.g., the word/object-level and the frame-level attention).

\begin{figure*}[ht]
\centering
  \includegraphics[width=\textwidth]{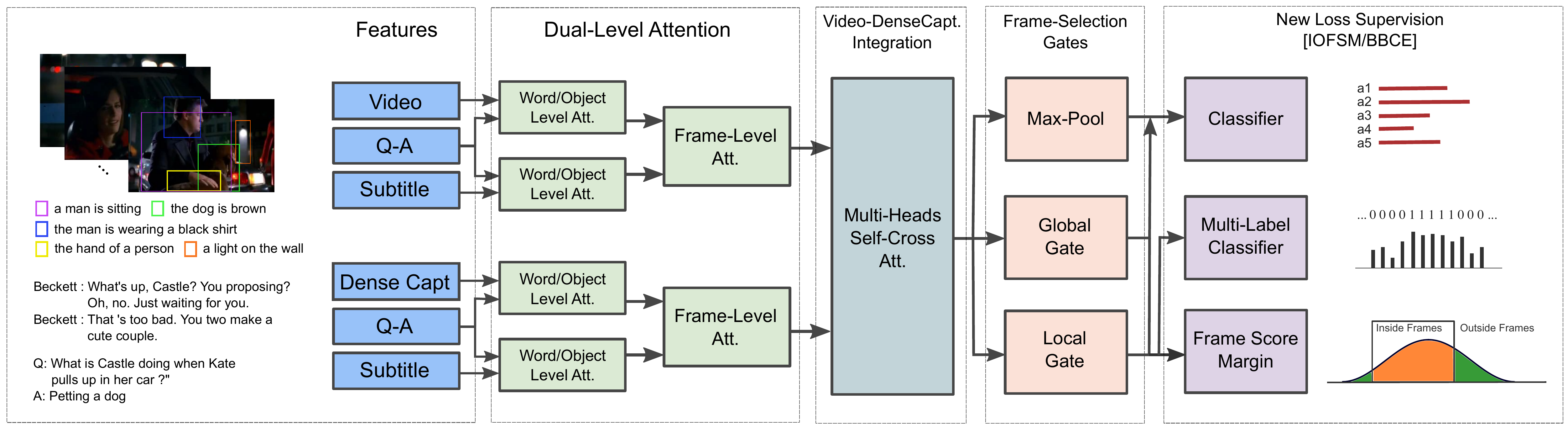}
  \vspace{-19pt}
\caption{Our model consists of three parts: Dual-Level Attention, Video-DenseCapt Integration, and Frame-Selection Gates. The new loss functions (IOFSM/BBCE) also help improve the model with enhanced supervision. \label{fig:model} \vspace{-15pt}}
\end{figure*}

\section{Related Work}
\paragraph{Visual/Video Question Answering}
Understanding visual information conditioned on language is an important ability for an agent who is supposed to have integrated intelligence. Many tasks have been proposed to evaluate such ability, and visual question answering is one of those tasks \cite{VQA, lu2016hierarchical, fukui2016multimodal, xu2016ask, yang2016stacked, zhu2016visual7w, balanced_vqa_v2, Anderson2017up-down}. Recently, beyond question answering on a single image, attention to understanding and extracting information from a sequence of images, i.e., a video, is rising \cite{tapaswi2016movieqa,maharaj2017dataset,kim2017deepstory,jang2017tgif, lei2018tvqa, zadeh2019social, lei2019tvqa+, garcia2020knowit}. Answering questions on videos requires an understanding of temporal information as well as spatial information, so it is more challenging than a single image question answering.

\paragraph{Temporal Localization}
Temporal localization is a task that is widely explored in event/object detection in video context. There has been work that solely processes visual information to detect objects/actions/activity \cite{gaidon2013temporal, weinzaepfel2015learning, shou2016temporal, dai2017temporal, shou2017cdc}. At the same time, work on natural language-related temporal localization task is less explored with recent work that focuses on the retrieval of a certain moment in a video by natural language \cite{anne2017localizing, gao2017tall}. With deliberately designed gating and attention mechanisms, our work, in general, will greatly contribute to the task of temporal localization, especially under natural language context and multimodal data.

\paragraph{Dense Image Captioning}
Image captioning is another direction of understanding visual and language information jointly. Single-sentence captions \cite{karpathy2015deep, Anderson2017up-down} capture the main concept of an image to describe it in a single sentence. However, an image could contain multiple aspects that are important/useful in different ways. Dense captions \cite{densecap, yang2017dense} and paragraph captions \cite{krause2017hierarchical,liang2017recurrent, melaskyriazi2018paragraph} have been introduced to densely and broadly capture the diverse aspects and salient regions of an image. Especially, dense caption describes an image in object level and gives useful salient regional information about objects such as attributes and actions. In this paper, we take advantage of this dense caption's ability to help our video QA model understand an image better for answering questions.

\section{Model}
Our model consists of 2 parts: feature fusion and frame selection. For feature fusion, we introduce dual-level (word/object and frame level) attention, and we design the frame selection problem as a multi-label classification task and introduce 2 new loss functions for enhanced supervision (Figure \ref{fig:model}).

\subsection{Features}
\begin{figure}[t]
\centering
  \includegraphics[width=.65\columnwidth]{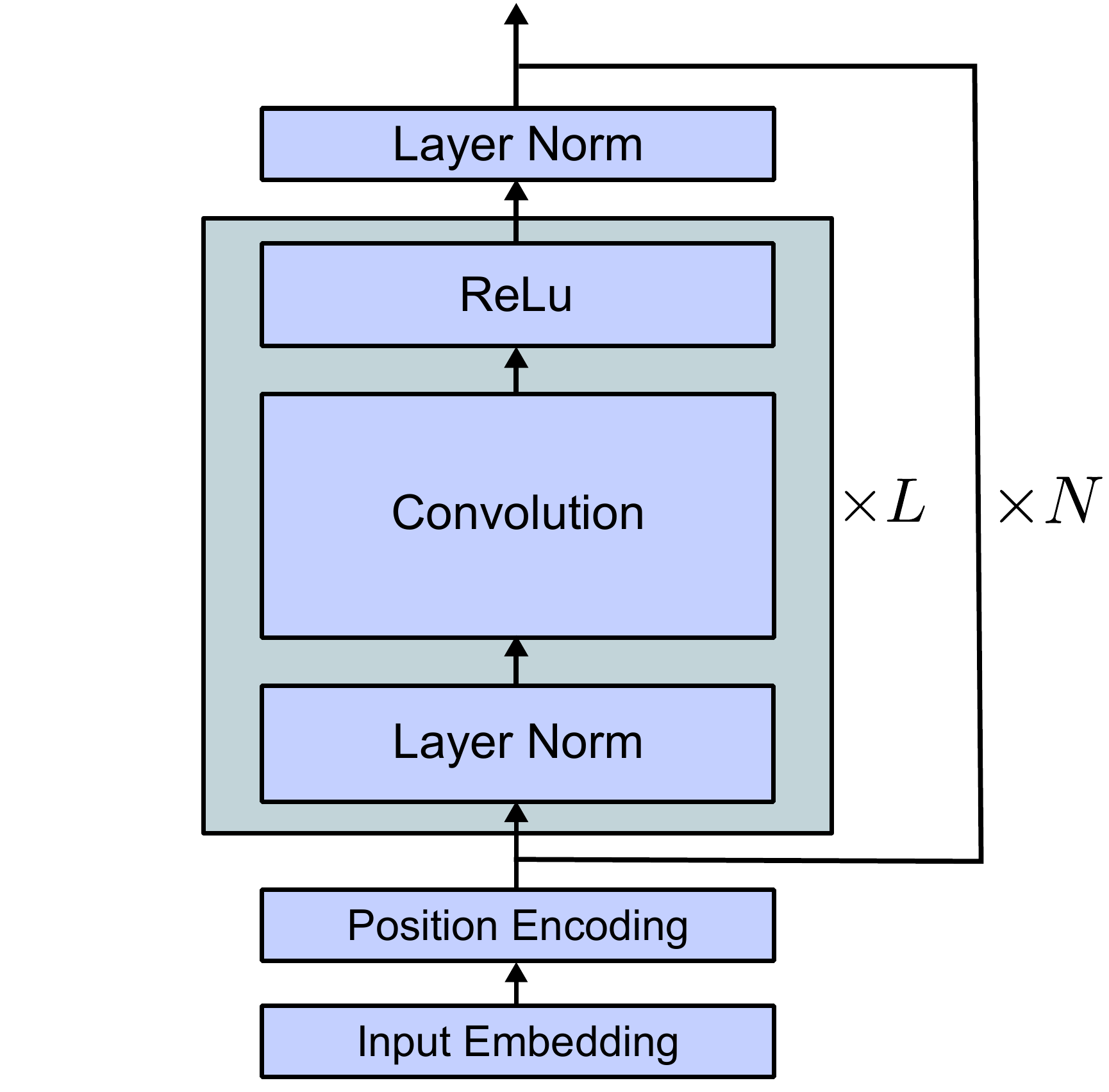}
    \vspace{-5pt}
\caption{CNN encoder. We use this block to encode all the input features. \label{fig:encoder}}
  \vspace{-10pt}
\end{figure}

We follow the same approach of \citet{lei2019tvqa+}'s work to obtain features from video, question-answer pairs, and subtitle input and encode them. We sample frames at 0.5 fps and extract object features from each frame via Faster R-CNN \cite{girshick2015fast}. Then we use PCA to get features of 300 dimension from top-20 object proposals. We also create five hypotheses by concatenating a question feature with each of five answer features, and we pair each visual frame feature with temporally neighboring subtitles. We encode all the features using convolutional encoder.
\begin{equation}
 \phi_{en}(x):\left\{
\begin{aligned}
x_{0}^0 &= E_{pos}(x)\\
x_{t}^i & = f_{i,t}(x_{t-1}^i) + x_{t-1}^i\text{, }\\
f_i(x_0^i) &= g_n(x_L^i)\\
y &= f_N \circ... \circ f_1 (x_0^0) \\
\end{aligned} 
\right. 
\end{equation}
where $E_{pos}$ denotes positional encoding, $f_{i,t}$ convolution preceded by Layer Normalization and followed by ReLU activation, and $g_n$ the layer normalization. The encoder is composed of \(N\) blocks iterations. In each iteration, the encoded inputs are transformed \(L\) times of convolutions. The \(L\) is set to 2, and \(N\) to 1 in our experiment (Figure \ref{fig:encoder}).

\subsection{Dual-Level Attention}
\begin{figure}[t]
\centering
  \includegraphics[width=.95\columnwidth]{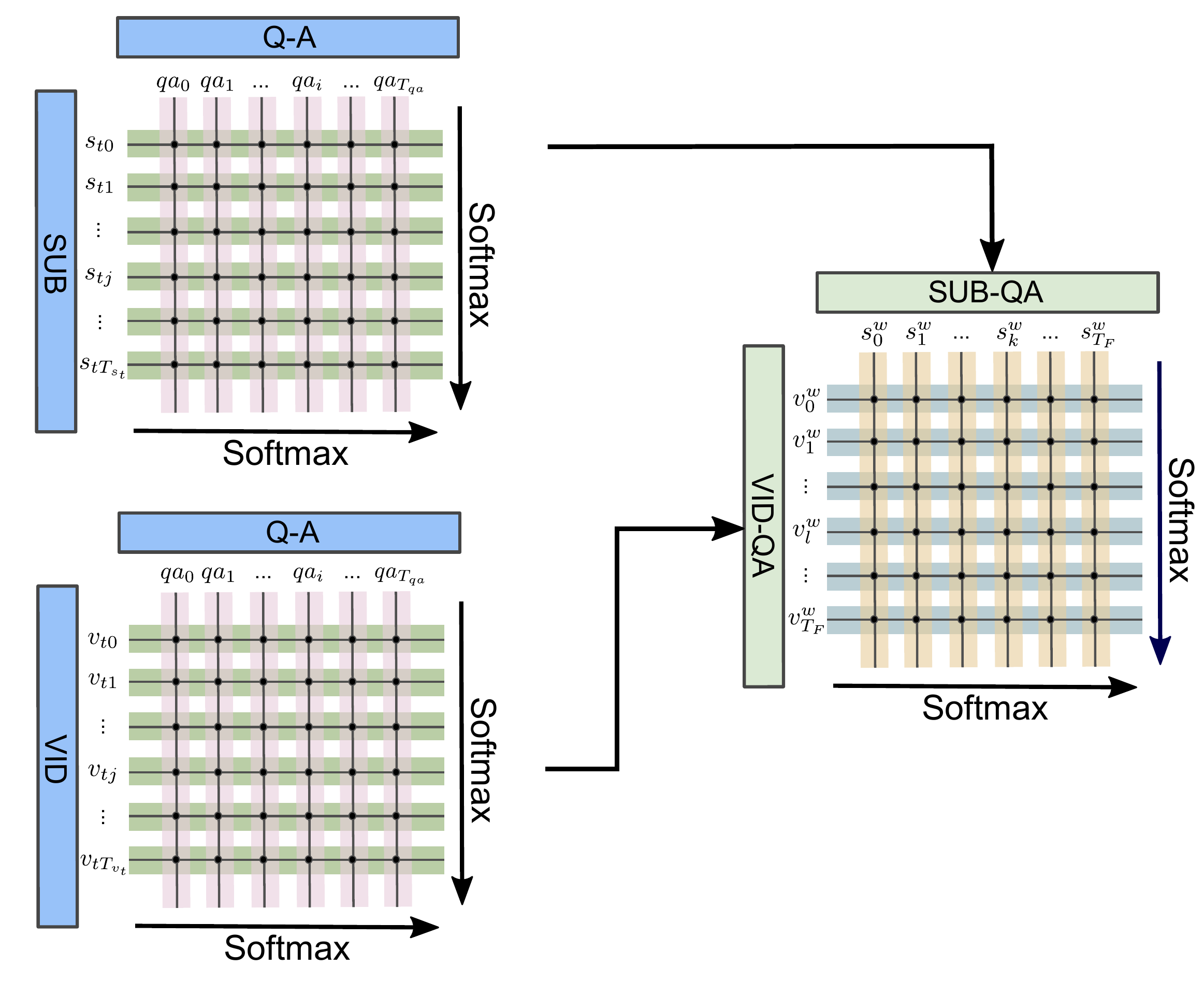}
  \vspace{-5pt}
\caption{Dual-Level Attention. Our model performs two-level attentions (word/object and frame level) sequentially. In the word/object-level attention, each word/object is aligned to relevant words or objects. In the frame-level attention, each frame (which has integrated information from the word/object-level attention) is aligned to relevant frames.
 \label{fig:cross_att}}
\vspace{-10pt}
\end{figure}

In dual-level attention, features are sequentially aligned in word/object-level and frame-level (Figure \ref{fig:cross_att}). 
\paragraph{Word/Object-Level Attention}
The QA feature, \(qa=\{qa_0, qa_1, .., qa_{T_{qa}}\}\), are combined with subtitle feature, \(s_t=\{s_{t0}, s_{t1}, .., s_{tT_{s_t}}\}\), and visual feature, \(v_t=\{v_{t0}, v_{t1}, .., v_{tT_{v_t}}\}\), of \(t\)-th frame respectively via word/object-level attention. To be specific, we calculate similarity matrices following \citet{seo2016bidirectional}'s approach, \(S_t^v\in \mathbb{R}^{T_{qa}\times T_{s_t}}\) and \(S_t^s \in \mathbb{R}^{T_{qa}\times T_{v_t}}\), from QA/subtitle and QA/visual features respectively. From the similarity matrices, attended subtitle features are obtained and combined with the QA features by concatenating and applying a transforming function. Then, max-pooling operation is applied word-wise to reduce the dimension.
\begin{align}
    (S_t^s)_{ij} &= qa_i^{\top}s_{tj}\\
    s_t^{att} &= \textrm{softmax}(S^s_t)\cdot s_t\\
    qa_s^{m}  &= \textrm{maxpool}(f_1([qa;s_t^{att};qa \odot s_t^{att}]))
\end{align}
where \(f_1\) is a fully-connected layer followed by \(\textrm{ReLU}\) non-linearity. The same process is applied to the QA features.
\begin{align}
    qa^{att} &= \textrm{softmax}(S_t^{s\top}) \cdot qa\\
    s_{t}^{m} &= \textrm{maxpool}(f_1([s_{t};qa^{att};s_{t} \odot qa^{att}]))
\end{align}
The fused features from different directions are integrated by concatenating and being fed to a function as follows:  
\begin{align}
    s_t^{w} &= f_2([qa_s^{m}; s_{t}^{m};qa_s^{m} \odot s_{t}^{m}; qa_s^{m} + s_{t}^{m}])
\end{align}
where \(f_2\) is the same function as \(f_1\) with non-shared parameters. All this process is also applied to visual features to get word/object-level attended features.
\begin{align}
    v_t^{w} &= f_2([qa_v^{m}; v_{t}^{m};qa_v^{m} \odot v_{t}^{m};qa_v^{m} + v_{t}^{m}])
\end{align}

\paragraph{Frame-Level Attention}
The fused features from word/object-level attention are integrated frame-wise via frame-level attention. Similar to the word/object-level attention, a similarity matrix, \(S \in \mathbb{R}^{T_{F}\times T_{F}}\),  is calculated, where \(T_F\) is the number of frames. Also, from the similarity matrix, attended frame-level features are calculated.
\begin{align}
    (S)_{kl} &= s_k^{w\top}v_l^{w}\\
    s^{att} &= \textrm{softmax}(S)\cdot s^w + s^w\\
    \hat{v} &= f_3([v^w;s^{att};v^w \odot s^{att}; v^w + s^{att}])
\end{align}
\begin{align}
    v^{att} &= \textrm{softmax}(S^{\top})\cdot v^w + v^w\\
    \hat{s} &= f_3([s^w;v^{att};s^w \odot v^{att}; s^w + v^{att}])
\end{align}
where \(f_3\) is the same function as \(f_1\) and \(f_2\) with non-shared parameters.
The frame-wise attended features are added to get an integrated feature.
\begin{align}
    u^{sv} &= \hat{s} + \hat{v}
\end{align}

\subsection{Video and Dense Caption Integration}
We also employ dense captions to help further improve the temporal localization of our video-QA model. They provide more diverse salient regional information (than the usual single non-dense image captions) about object-level details of image frames in a video clip, and also allow the model to explicitly (in textual/semi-symbolic form) match keywords/patterns between dense captions and questions to find relevant locations/frames.

\begin{figure}[t]
\centering
  \includegraphics[width=.75\columnwidth]{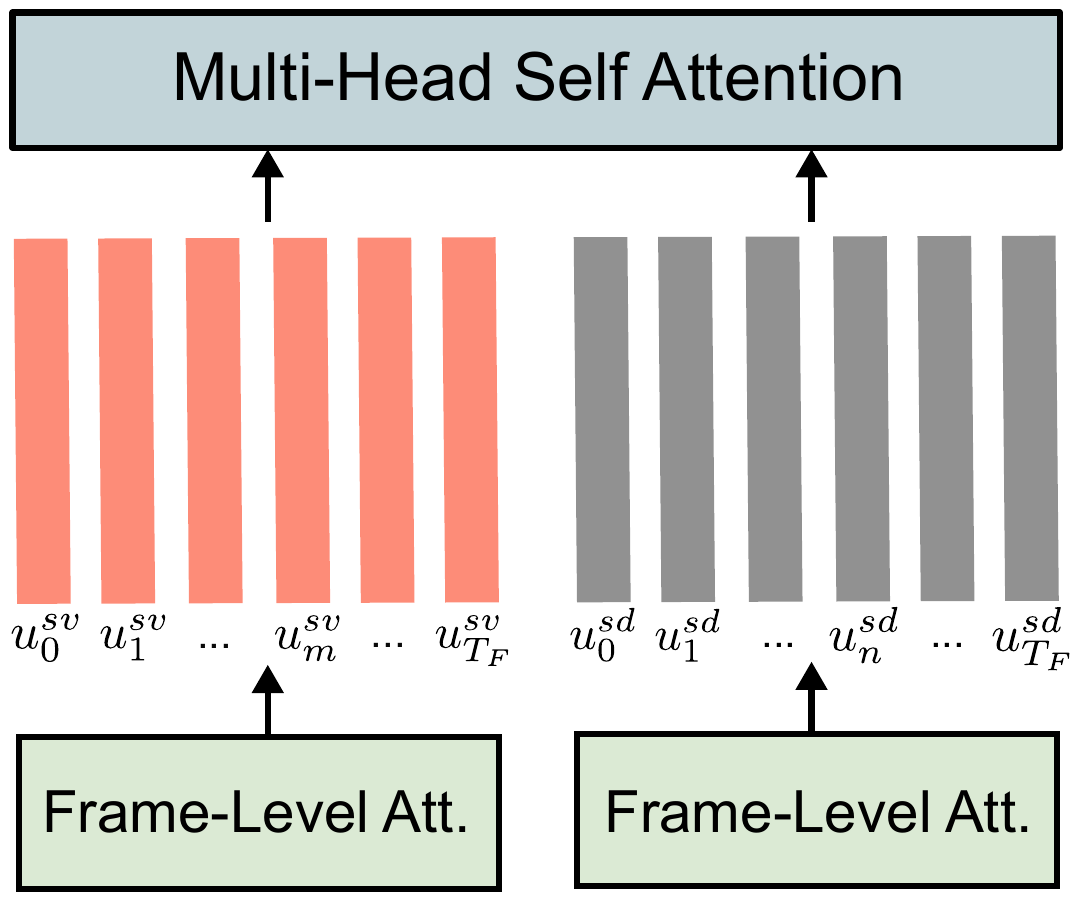}
  \vspace{-5pt}
\caption{Self-Cross Attention. We combine information each from the video (fused with subtitle and QA) and dense caption (fused with subtitle and QA) via the multi-head self attention. Before being fed to the multi-head self attention module, video and dense caption features are concatenated. Thus, self and cross attentions are performed simultaneously. \label{fig:self_cross}}
\vspace{-10pt}
\end{figure}

We apply the same procedure to the dense caption feature by substituting video features with dense caption features to obtain \(u^{sd}\). To integrate \(u^{sv}\) and \(u^{sd}\), we employ multi-head self attention (Figure \ref{fig:self_cross}). To be specific, we concatenate \(u^{sv}\) and \(u^{sd}\) frame-wise then feed them to the self attention function. 
\begin{equation}
\phi_{\text{self-att}}(x)\left\{
\begin{aligned}
h_{\mathrm{i}} &= g_{a}(w_q^{\top}x_i , w_k^{\top}x_i, w_v^{\top}x_i)\\
y &=w_{m}^{\top}[h_{1}; \ldots;h_{\mathrm{k}}] 
\end{aligned}
\right.
\end{equation}
where $g_{a}$ denotes self-attention.
\begin{align}
    u^{svd} = \phi_{\text{self-att}}([u^{sv};u^{sd}])
\end{align}
In this way, \(u^{sv}\) and \(u^{sd}\) attend to themselves while attending to each other simultaneously. We split the output, \(u^{svd}\) into the same shape as the input, then add the two. 
\begin{align}
    z &= u^{svd}[0:T_F] + u^{svd}[T_F:2T_F]
\end{align}

\subsection{Frame-Selection Gates}
To select appropriate information from the frame-length features, we employ max-pooling and gates. Features from the video-dense caption integration are fed to the CNN encoder. A fully-connected layer and sigmoid function are applied sequentially to the output feature to get frame scores that indicate how relevant each frame is for answering a given question. We get weighted features by multiplying the output feature from the CNN encoder with the scores.
\begin{align}
    \hat{z} &= \phi_{\text{en2}}(z)\\
    g^{L} &= \textrm{sigmoid}(f^{L}(\hat{z})) \\
    z^{gl} &= \hat{z} \odot g^{L}
\end{align}
We calculate another frame scores with a different function \(f^G\) to get another weighted feature.
\begin{align}
 g^{G} &= \textrm{sigmoid}(f^{G}(\hat{z})) \\
    z^{gg} &= \hat{z} \odot g^{G}
\end{align}
Finally, following \citet{lei2019tvqa+}'s work, we also apply frame-wise max-pooling.
\begin{align}
    z^{max} &= \textrm{maxpool}(\hat{z})
\end{align}
The three features (from local gate, global gate, and max-pooling, respectively), are then concatenated and fed to the classifier to give scores for each candidate answer.
\begin{align}
    logit &= \textrm{clssifier}([z^{max};z^{gg};z^{gl}])
\end{align}
We get the logits for the five candidate answers and choose the highest value as the predicted answer.
\begin{align}
    loss_{cls} &=-\textrm{log}(\frac{e^{s_g}}{\sum_ke^{s_k}})
\end{align}
where \(s_g\) is the logit of ground-truth answer.

\subsection{Novel Frame-Selection Supervision Loss Functions}
We cast frame selection as a multi-label classification task. The frame scores from the local gate, \(g^L\), are supervised by human importance annotations, which are time spans (start-end points pair) annotators think needed for selecting correct answers. To this end, we transform the time span into ground-truth frame scores, i.e., if a frame is within the time span, the frame has `1' as its label and a frame outside the span gets `0'. In this way, we can assign a label to each frame, and frames should get as close scores as their ground-truth labels. We train the local gate network with binary cross-entropy (BCE) loss.
\begin{align}
    loss_{bce} &=- \sum^{T_F}_i(y\textrm{log}(s_i^f) + (1-y)\textrm{log}(1-s_i^f))
\end{align}
where \(s_i^f\) is a frame score of \(i\)-th frame, and y is a corresponding ground-truth label.

\paragraph{In-and-Out Frame Score Margin}
For additional supervision other than the binary cross-entropy loss, we create a novel loss function, In-and-Out Frame Score Margin (IOFSM). 
\begin{align}
    loss_{io} &= 1 + \textrm{Avg}(\textrm{OFS}) - \textrm{Avg}(\textrm{IFS})
\end{align}
where OFS (Out Frame Score) is scores of frames whose labels are `0' and IFS (In Frame Score) is scores of frames whose labels are `1'.

\begin{savenotes}
\begin{table*}[t]
% \small
\begin{center}
\resizebox{1.95\columnwidth}{!}{
 \begin{tabular}{|c|l|c||c|c|c|c|c|c|c|}
  \hline
 &  \multirow{2}{*}{Model} & \multicolumn{7}{c|}{Test-Public (\%)} & \multirow{2}{*}{Val (\%)} \\
  \cline{3-9} 
 && all  & bbt & friends & himym &grey & house & castle &\\
 \hline
    1 & jacobssy (anonymous) & 66.01 & 68.75 & 64.98 & 65.08 & 69.22 & 66.45 & 63.74 & 64.90\\
    2 & multi-stream \cite{lei2018tvqa} & 66.46 & 70.25 & 65.78 & 64.02 & 67.20 & 66.84 & 63.96 & 65.85\\
    3 & PAMN \cite{kim2019progressive} & 66.77 & - & - & - & - & - & - & 66.38\\
    4 & Multi-task \cite{kim2019gaining} & 67.05 & - & - & - & - & - & - & 66.22\\
    5 & ZGF (anonymous) & 68.77 & - & - & - & - & - & - & 68.90\\
    6 & STAGE \cite{lei2019tvqa+} & 70.23 & - & - & - & - & - & - & 70.50\\
    7 & akalsdnr (anonymous) & 70.52 & 71.49 & 67.43 & 72.22 & 70.42 & 70.83 & 72.30 & 71.13 \\
    \hline
    8 & Ours (hstar) & \textbf{74.09} &\textbf{74.04}& \textbf{73.03}& \textbf{74.34} & \textbf{73.44} & \textbf{74.68} & \textbf{74.86} & \textbf{74.20}\\
   \hline
\end{tabular}
}
\end{center}
\vspace{-10pt}
\caption{Our model outperforms the state-of-the-art models by a large margin. Moreover, the scores of our model across all the TV shows are more balanced than the scores from other models, which means our model is more consistent/robust and not biased to the dataset from specific TV shows.\footnote{At the time of the ACL2020 submission deadline, the publicly visible rank-1 entry was 70.52\%. Since then, two more entries have appeared in the leaderboard; however, our method still outperforms their scores by a large margin (71.48\% and 71.13\% versus 74.09\%).} \label{tbl:main_result}
}
\vspace{-10pt}
\end{table*}
\end{savenotes}

\paragraph{Balanced Binary Cross-Entropy}
In our multi-label classification setting, each frame can be considered as one training example. Thus, the total number of examples and the proportion between positive and negative examples vary for every instance. This variation can cause unbalanced training since negative examples usually dominate. To balance the unbalanced training, we apply a simple but effective modification to the original BCE, and we call it Balanced Binary Cross-Entropy (BBCE). To be specific, instead of summing or averaging through the entire frame examples, we divide the positive and negative examples and calculate the average cross-entropy scores separately, then sum them together.
\begin{align}
\begin{split}
    loss_{bbce} &=- \Big( \sum^{T_{F_{in}}}_i\textrm{log}(s_i^{f_{in}}) / T_{F_{in}} \\
    &+ \sum^{T_{F_{out}}}_j\textrm{log}(1-s_j^{f_{out}})/T_{F_{out}}\Big)
\end{split}
\end{align}
where \(s_i^{f_{in}}\) and \(s_j^{f_{out}}\) are \(i\)-th in-frame score and \(j\)-th out-frame score respectively, and \(T_{F_{in}}\) and \(T_{F_{out}}\) are the number of in-frames and out-frames respectively.

Thus, the total loss is:
\begin{align}
    loss &= loss_{cls} + loss_{(b)bce} + loss_{io}
\end{align}

\section{Experimental Setup}
\paragraph{TVQA Dataset}
TVQA dataset \cite{lei2018tvqa} consists of video frames, subtitles, and question-answer pairs from 6 TV shows. The number of examples for train/validation/test-public dataset are 122,039/15,253/7,623. Each example has five candidate answers with one of them the ground-truth. So, TVQA is a classification task, in which models select one from the five candidate answers, and models can be evaluated on the accuracy metric.
\paragraph{Dense Captions}
We use \citet{yang2017dense}'s pretrained model to extract dense captions from each video frame. We extract the dense captions in advance and use them as extra input data to the model.\footnote{This is less computationally expensive and dense captions from the separately trained model will be less biased towards the questions of TVQA dataset, and hence provide more diverse aspects of image frames of a video clip.}

\paragraph{Training Details}
We use GloVe \cite{pennington2014glove} word vectors with dimension size of 300 and RoBERTa \cite{liu2019roberta} with 768 dimension. The dimension of the visual feature is 300, and the base hidden size of the whole model is 128. We use Adam \cite{kingma2014adam} as the optimizer. We set the initial learning rate to 0.001 and drop it to 0.0002 after running 10 epochs. For dropout, we use the probability of 0.1.

\section{Results and Ablation Analysis}
As seen from Table \ref{tbl:main_result}, our model outperforms the state-of-the-art models in the TVQA leaderboard. Especially our model gets balanced scores for all the TV shows while some other models have high variances across the shows. As seen from Table \ref{tbl:score_std}, the standard deviation and `max-min' value over our model's scores for each TV show are 0.65 and 1.83, respectively, which are the lowest values among all models in the list. This low variance could mean that our model is more consistent and robust across all the TV shows.

\paragraph{Model Ablations}

As shown in Table \ref{tbl:model_ablation}, our basic dual-attention and frame selection gates model shows substantial improvement over the strong single attention and frame span baseline (row 4 vs 1: $p < 0.0001$), which is from the best published model \cite{lei2019tvqa+}. Each of our dual-attention and frame selection gates alone shows a small improvement in performance than the baseline (row 3 vs 1 and 2 vs 1, respectively).\footnote{Although the improvements are not much, but performing word/object level attention and then frame level attention is more intuitive and interpretable than a non-dual-attention method, allowing us to show how the model works: see visualization in Sec.~\ref{sec:vis}.} However, when they are applied together, the model works much better. The reason why they are more effective when put together is that frame selection gates basically select frames based on useful information from each frame feature and our dual-attention can help this selection by getting more relevant information to each frame through the frame-level attention. Next, our new loss functions significantly help over the dual-attention and frame selection gates model by providing enhanced supervision (row 5 vs 4: $p < 0.0001$, row 7 vs 6: $p < 0.005$). Our RoBERTa version is also significantly better than the GloVe model (row 6 vs 4: $p <0.0005$, row 7 vs 5: $p < 0.01$). Finally, employing dense captions further improves the performance via useful textual clue/keyword matching (row 8 vs 7: $p < 0.005$).\footnote{Statistical significance is computed using the bootstrap test~\cite{efron1994introduction}.}

\begin{savenotes}
\begin{table}[t]
% \small
\begin{center}
\resizebox{0.95\columnwidth}{!}{
 \begin{tabular}{|c|l|c|c|c|}
  \hline
 & \multirow{2}{*}{Model} & \multicolumn{3}{c|}{TV Show Score}\\
  \cline{3-5}
 && avg. & std. &max-min\\
 \hline
 1 & jacobssy (anonymous) & 66.37 & 2.01& 5.48\\
    2 & multi-stream \cite{lei2018tvqa} & 66.34 &2.15 & 6.29 \\
    3 & akalsdnr (anonymous) & 70.78 & 1.65 & 4.87 \\
    \hline
    4 & Ours & 74.07 & 0.65 & 1.83  \\
   \hline
\end{tabular}
}
\end{center}
\vspace{-5pt}
\caption{Average and standard deviation of the test-public scores from each TV show (for this comparison, we only consider models that release the scores for each TV show).\footnote{Two more entries have appeared in the leaderboard since the ACL2020 submission deadline. However, our scores are still more balanced than their scores across all TV shows (std.: 2.11 and 2.40 versus our 0.65, max-min: 5.50 and 7.38 versus our 1.83).} \label{tbl:score_std}
} 
\vspace{-5pt}
\end{table}
\end{savenotes}

\begin{table}[t]
% \small
\begin{center}
\resizebox{0.95\columnwidth}{!}{
 \begin{tabular}{|c|l|c|}
  \hline
 & Model & Val Score (\%)\\
 \hline
 1 & Single-Att + Frame-Span & 69.86\\
 2 & Single-Att + Frame-Selection Gates & 70.08\\
 3 & Dual-Att + Frame-Span & 70.20\\
 4 & Dual-Att + Frame-Selection Gates (w/o NewLoss) & 71.26 \\
 5 & Dual-Att + Frame-Selection Gates & 72.51  \\
 6 & Dual-Att + Frame-Selection Gates (w/o NewLoss) + RoBERTa &  72.53  \\
 7 & Dual-Att + Frame-Selection Gates + RoBERTa &  73.34  \\
 8 & Dual-Att + Frame-Selection Gates + RoBERTa + DenseCapts & 74.20  \\
   \hline
\end{tabular}
}
\end{center}
\vspace{-5pt}
\caption{Model Ablation: our dual-attention / frame-selection Gates, new loss functions, and dense captions help improve the model's performance (NewLoss: IOFSM+BBCE).\label{tbl:model_ablation}
} 
\vspace{-10pt}
\end{table}

\begin{table}[t]
% \small
\begin{center}
\resizebox{0.95\columnwidth}{!}{
 \begin{tabular}{|c|l|c|c|c|c|c|}
  \hline
 & \multirow{2}{*}{Loss} &  \multirow{2}{*}{Val Score (\%)}& \multicolumn{2}{c|}{IFS} & \multicolumn{2}{c|}{OFS} \\
 \cline{4-7} 
 &&&avg&std&avg&std\\
 \hline
    1 & BCE & 71.26 & 0.468 & 0.108 & 0.103 & 0.120 \\
    2 & IOFSM & 70.75 & 0.739 & 0.127 & 0.143 &0.298\\
    3 & BCE+IOFSM & 72.22 & 0.593 & 0.128& 0.111 & 0.159 \\
    4 & BBCE & 72.27 & 0.759 &  0.089 & 0.230 & 0.231 \\
    5 & BBCE+IOFSM & 72.51 & 0.764 &  0.098 & 0.182 & 0.246 \\
   \hline
\end{tabular}
}
\end{center}
\vspace{-5pt}
\caption{IOFSM and BBCE help improve the model's performance by changing in and out-frame scores. \label{tbl:bce_iofsm}
} 
\vspace{-10pt}
\end{table}

\paragraph{IOFSM and BCE Loss Functions Ablation and Analysis}
To see how In-and-Out Frame Score Margin (IOFSM) and Binary Cross-Entropy (BCE) loss affect the frame selection task, we compare the model's performance/behaviors according to the combination of IOFSM and BCE. As shown in Table \ref{tbl:bce_iofsm}, applying IOFSM on top of BCE gives a better result. 
When we compare row 1 and 3 in Table \ref{tbl:bce_iofsm}, the average in-frame score of BCE+IOFSM is higher than BCE's while the average out-frame scores of both are almost the same. This can mean two things: (1) IOFSM helps increase the scores of in-frames, and (2) increased in-frame scores help improve the model's performance. On the other hand, when we compare row 1 and 2, the average in-frame score of IOFSM is higher than BCE's. But, the average out-frame score of IOFSM is also much higher than BCE's. This can mean that out-frame scores have a large impact on the performance as well as in-frame scores. This is intuitively reasonable. Because information from out-frames also flows to the next layer (i.e., classifier) after being multiplied by the frame scores, the score for the `negative' label also has a direct impact on the performance. So, making the scores as small as possible is also important. Also, when we compare the row 2 and others (2 vs. 1 and 3), the gap between in-frame scores is much larger than the gap between out-frame scores. But, considering the scores are average values, and the number of out-frames is usually much larger than in-frames, the difference between out-frame scores would affect more than the gap itself.

\paragraph{Balanced BCE Analysis}
We can see from row 1 and 4 of the Table \ref{tbl:bce_iofsm} that BBCE shift the average scores of both in-frames and out-frames to higher values. This can show that scores from the BCE loss are biased to the negative examples, and BBCE can adjust the bias with the separate averaging. The score shift can help improve the model's performance. But, when comparing row 2 and 4, the out-frame scores of BBCE are higher than IOFSM, and this may imply that the result from BBCE should be worse than IOFSM since out-frame scores have a large impact on the performance. However, as we can see from row 2, the standard deviation of IOFSM's out-frame scores is larger than BBCE. This could mean that a model with IOFSM has an unstable scoring behavior, and it could affect the performance. As seen from row 5, applying BBCE and IOFSM together gives further improvement, possibly due to the increased in-frame scores and decreased out-frame scores while staying around at a similar standard deviation value.

\section{Visualizations \label{sec:vis}}

In this section, we visualize the dual-level attention (word/object and frame level) and the frame score change by new losses application (for all these attention examples, our model predicts the correct answers).

\begin{figure}[t]
\centering
  \includegraphics[width=.95\columnwidth]{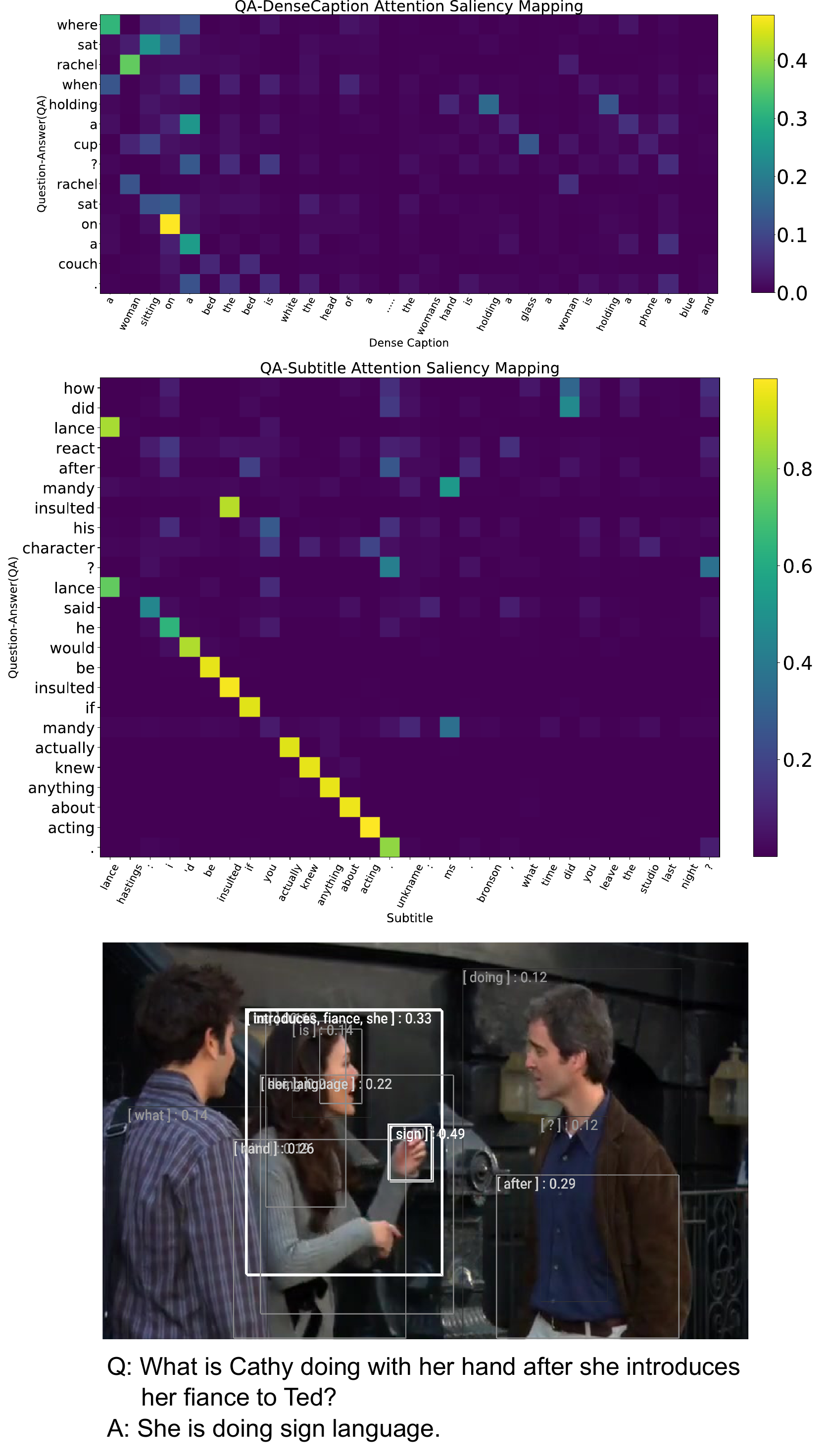}
  \vspace{-3pt}
\caption{Visualization of word/object level attention. Top: words from a question-answer pair to words from dense captions alignment. Middle: words from a question-answer pair to words from subtitles alignment. Bottom: words from a question-answer pair to regions (boxes) from an image (only boxes with top 1 scores from each word are shown).  \label{fig:word_att}}
\vspace{-3pt}
\end{figure}

\begin{figure}[t]
\centering
  \includegraphics[width=0.8\columnwidth]{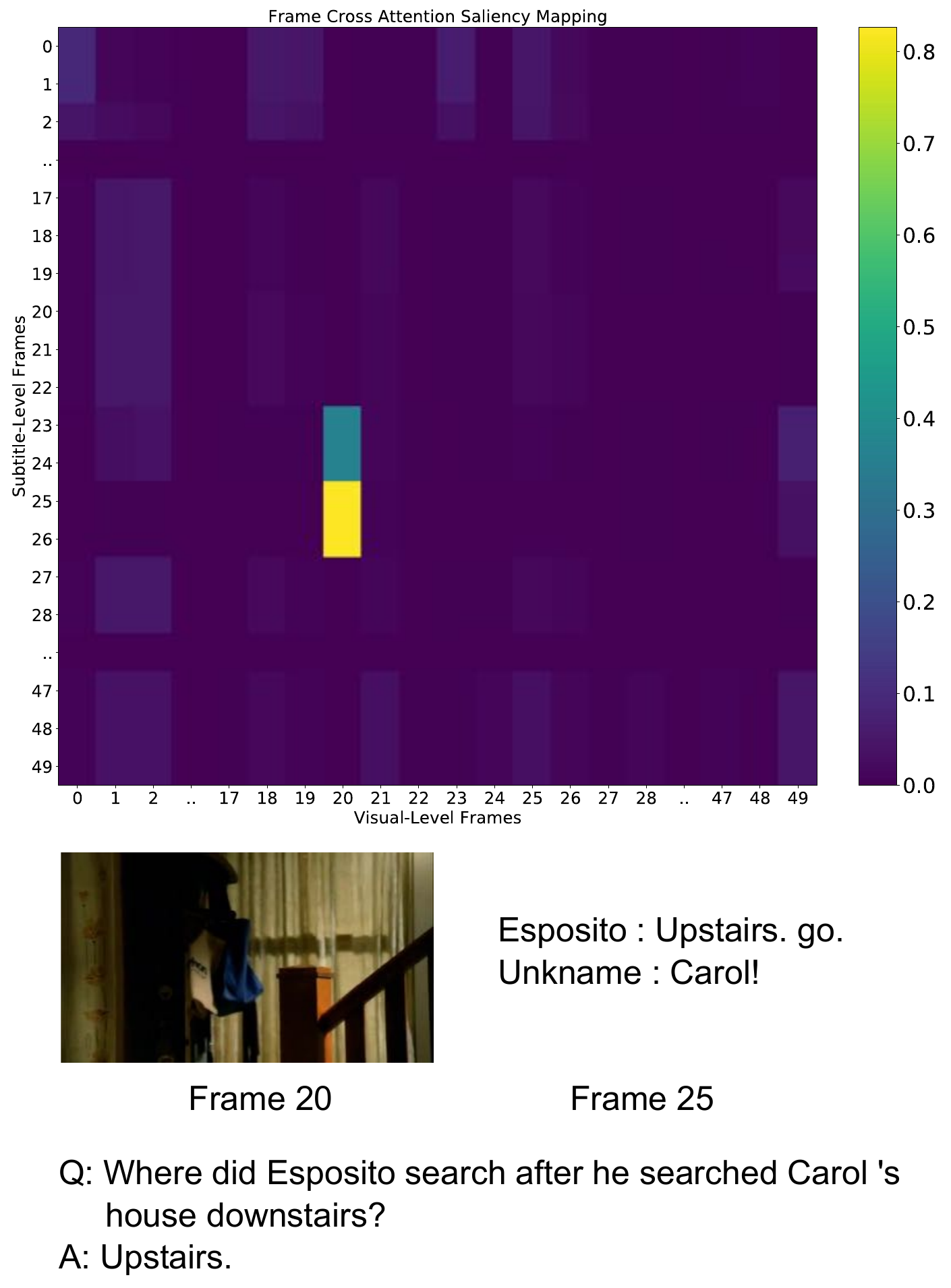}
  \vspace{-2pt}
\caption{Visualization of frame-level attention. Frame 25 (which contains `upstairs') from subtitle features and frame 20 (which shows `downstairs' by banister upward) from visual features are aligned.\label{fig:frame_level_att}}
\vspace{-1pt}
\end{figure}

\begin{figure}[t]
\centering
  \includegraphics[width=0.95\columnwidth]{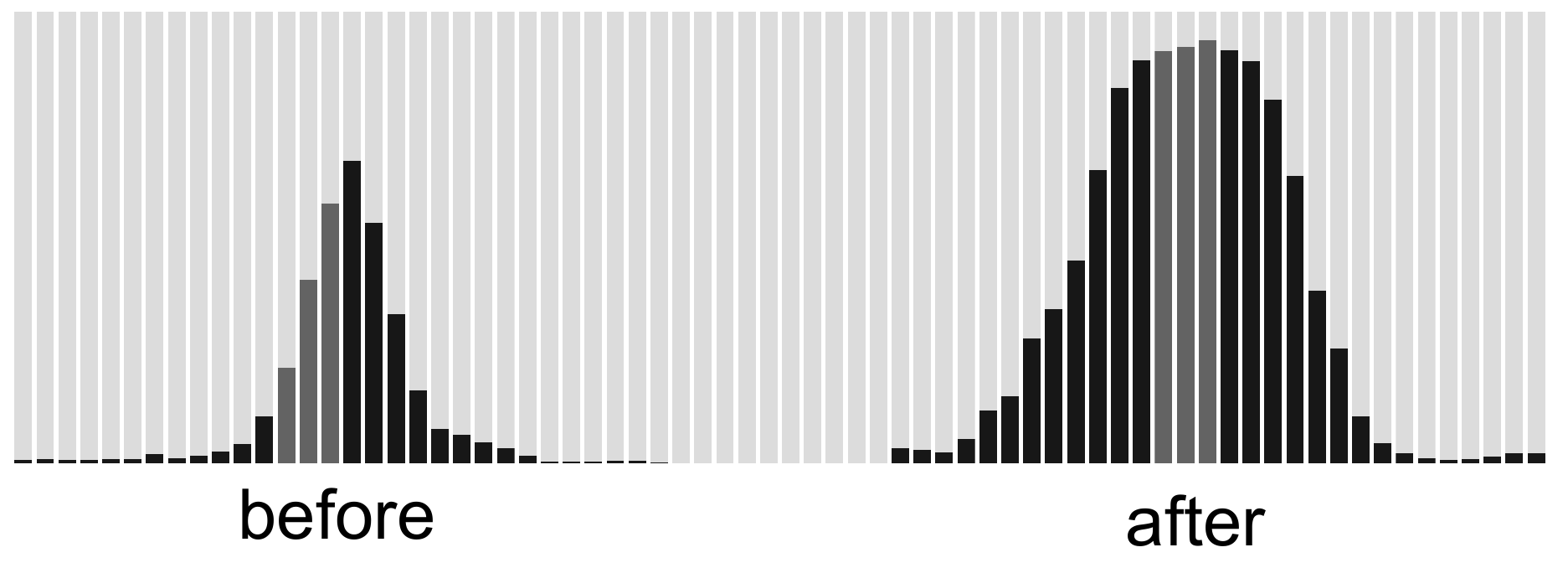}
  \vspace{-5pt}
\caption{Visualization of distribution change in frame selection scores. Left: the score distribution before applying new losses (IOFSM+BBEC). Right: the score distribution after applying the losses. Scores neighboring in-frame (gray) are increased. For this example, the model does not predict the right answer before applying the losses, but after training with the losses, the model chooses the correct answer. \label{fig:frame_select}}
\vspace{-3pt}
\end{figure}

\paragraph{Word/Object-Level Attention}
We visualize word-level attention in Figure \ref{fig:word_att}. In the top example, the question and answer pair is ``Where sat Rachel when holding a cup?'' - ``Rachel sat on a couch''. Our word/object-level attention between QA pair and dense caption attend to a relevant description like `holding a glass' to help answer the question. In the middle example, the question and answer pair is, ``How did Lance react after Mandy insulted his character?'' - ``Lance said he would be insulted if Mandy actually knew anything about acting''. Our word/object-level attention between QA pair and subtitle properly attend to the most relevant words such as `insulted', `knew', and `acting' to answer the question. 
In the bottom example, the question and answer pair is, ``What is Cathy doing with her hand after she introduces her fiance to Ted?'' - ``She is doing sign language''. From the score of our word/object-level attention, the model aligns the word `sign' to the woman's hand to answer the question.

\paragraph{Frame-Level Attention}
As shown in Figure \ref{fig:frame_level_att}, our frame-level attention can align relevant frames from different features. In the example, the question and answer pair is ``Where did Esposito search after he searched Carol's house downstairs?'' - ``Upstairs''. To answer this question, the model needs to find a frame in which `he (Esposito) searched Carol's house downstairs', then find a frame which has a clue for `where did Esposito search'. Our frame-level attention can properly align the information fragments from different features (Frame 20 and 25) to help answer questions.

\paragraph{Frame Score Enhancement by New Losses}
As seen in Figure \ref{fig:frame_select}, applying our new losses (IOFSM+BBCE) changes the score distribution over all frames. Before applying our losses (left figure), overall scores are relatively low. After using the losses, overall scores increased, and especially, scores around in-frames get much higher. 
\section{Conclusion}
We presented our dual-level attention and frame-selection gates model and novel losses for more effective frame-selection. Furthermore, we employed dense captions to help the model better find clues from salient regions for answering questions. Each component added to our base model architecture (proposed loss functions and the adoption of dense captions) significantly improves the model's performance. Overall, our model outperforms the state-of-the-art models on the TVQA leaderboard, while showing more balanced scores on the diverse TV show genres.

\section*{Acknowledgments}
We thank the reviewers for their helpful comments. This work was supported by NSF Award 1840131, ARO-YIP Award W911NF-18-1-0336, DARPA KAIROS Grant FA8750-19-2-1004, and awards from Google and Facebook. The views, opinions, and/or findings contained in this article are those of the authors and should not be interpreted as representing the official views or policies, either expressed or implied, of the funding agency.

\bibliography{acl2020}

\begin{thebibliography}{40}
\expandafter\ifx\csname natexlab\endcsname\relax\def\natexlab#1{#1}\fi

\bibitem[{Anderson et~al.(2018)Anderson, He, Buehler, Teney, Johnson, Gould,
  and Zhang}]{Anderson2017up-down}
Peter Anderson, Xiaodong He, Chris Buehler, Damien Teney, Mark Johnson, Stephen
  Gould, and Lei Zhang. 2018.
\newblock Bottom-up and top-down attention for image captioning and visual
  question answering.
\newblock In \emph{CVPR}.

\bibitem[{Anne~Hendricks et~al.(2017)Anne~Hendricks, Wang, Shechtman, Sivic,
  Darrell, and Russell}]{anne2017localizing}
Lisa Anne~Hendricks, Oliver Wang, Eli Shechtman, Josef Sivic, Trevor Darrell,
  and Bryan Russell. 2017.
\newblock Localizing moments in video with natural language.
\newblock In \emph{Proceedings of the IEEE International Conference on Computer
  Vision}, pages 5803--5812.

\bibitem[{Antol et~al.(2015)Antol, Agrawal, Lu, Mitchell, Batra, Zitnick, and
  Parikh}]{VQA}
Stanislaw Antol, Aishwarya Agrawal, Jiasen Lu, Margaret Mitchell, Dhruv Batra,
  C.~Lawrence Zitnick, and Devi Parikh. 2015.
\newblock {VQA}: {V}isual {Q}uestion {A}nswering.
\newblock In \emph{International Conference on Computer Vision (ICCV)}.

\bibitem[{Dai et~al.(2017)Dai, Singh, Zhang, Davis, and
  Qiu~Chen}]{dai2017temporal}
Xiyang Dai, Bharat Singh, Guyue Zhang, Larry~S Davis, and Yan Qiu~Chen. 2017.
\newblock Temporal context network for activity localization in videos.
\newblock In \emph{Proceedings of the IEEE International Conference on Computer
  Vision}, pages 5793--5802.

\bibitem[{Efron and Tibshirani(1994)}]{efron1994introduction}
Bradley Efron and Robert~J Tibshirani. 1994.
\newblock \emph{An introduction to the bootstrap}.
\newblock CRC press.

\bibitem[{Fukui et~al.(2016)Fukui, Park, Yang, Rohrbach, Darrell, and
  Rohrbach}]{fukui2016multimodal}
Akira Fukui, Dong~Huk Park, Daylen Yang, Anna Rohrbach, Trevor Darrell, and
  Marcus Rohrbach. 2016.
\newblock Multimodal compact bilinear pooling for visual question answering and
  visual grounding.
\newblock In \emph{Proceedings of the 2016 Conference on Empirical Methods in
  Natural Language Processing}, pages 457--468.

\bibitem[{Gaidon et~al.(2013)Gaidon, Harchaoui, and
  Schmid}]{gaidon2013temporal}
Adrien Gaidon, Zaid Harchaoui, and Cordelia Schmid. 2013.
\newblock Temporal localization of actions with actoms.
\newblock \emph{IEEE transactions on pattern analysis and machine
  intelligence}, 35(11):2782--2795.

\bibitem[{Gao et~al.(2017)Gao, Sun, Yang, and Nevatia}]{gao2017tall}
Jiyang Gao, Chen Sun, Zhenheng Yang, and Ram Nevatia. 2017.
\newblock Tall: Temporal activity localization via language query.
\newblock In \emph{Proceedings of the IEEE International Conference on Computer
  Vision}, pages 5267--5275.

\bibitem[{Garcia et~al.(2020)Garcia, Otani, Chu, and
  Nakashima}]{garcia2020knowit}
Noa Garcia, Mayu Otani, Chenhui Chu, and Yuta Nakashima. 2020.
\newblock Knowit vqa: Answering knowledge-based questions about videos.
\newblock In \emph{Proceedings of the Thirty-Fourth AAAI Conference on
  Artificial Intelligence}.

\bibitem[{Girshick(2015)}]{girshick2015fast}
Ross Girshick. 2015.
\newblock Fast r-cnn.
\newblock In \emph{Proceedings of the IEEE international conference on computer
  vision}, pages 1440--1448.

\bibitem[{Goyal et~al.(2017)Goyal, Khot, Summers{-}Stay, Batra, and
  Parikh}]{balanced_vqa_v2}
Yash Goyal, Tejas Khot, Douglas Summers{-}Stay, Dhruv Batra, and Devi Parikh.
  2017.
\newblock Making the {V} in {VQA} matter: Elevating the role of image
  understanding in {V}isual {Q}uestion {A}nswering.
\newblock In \emph{Conference on Computer Vision and Pattern Recognition
  (CVPR)}.

\bibitem[{Jang et~al.(2017)Jang, Song, Yu, Kim, and Kim}]{jang2017tgif}
Yunseok Jang, Yale Song, Youngjae Yu, Youngjin Kim, and Gunhee Kim. 2017.
\newblock Tgif-qa: Toward spatio-temporal reasoning in visual question
  answering.
\newblock In \emph{Proceedings of the IEEE Conference on Computer Vision and
  Pattern Recognition}, pages 2758--2766.

\bibitem[{Johnson et~al.(2016)Johnson, Karpathy, and Fei-Fei}]{densecap}
Justin Johnson, Andrej Karpathy, and Li~Fei-Fei. 2016.
\newblock Densecap: Fully convolutional localization networks for dense
  captioning.
\newblock In \emph{Proceedings of the IEEE Conference on Computer Vision and
  Pattern Recognition}.

\bibitem[{Karpathy and Fei-Fei(2015)}]{karpathy2015deep}
Andrej Karpathy and Li~Fei-Fei. 2015.
\newblock Deep visual-semantic alignments for generating image descriptions.
\newblock In \emph{Proceedings of the IEEE conference on computer vision and
  pattern recognition}, pages 3128--3137.

\bibitem[{Kim and Bansal(2019)}]{kim-bansal-2019-improving}
Hyounghun Kim and Mohit Bansal. 2019.
\newblock Improving visual question answering by referring to generated
  paragraph captions.
\newblock In \emph{Proceedings of the 57th Annual Meeting of the Association
  for Computational Linguistics}.

\bibitem[{Kim et~al.(2019{\natexlab{a}})Kim, Ma, Kim, Kim, and
  Yoo}]{kim2019gaining}
Junyeong Kim, Minuk Ma, Kyungsu Kim, Sungjin Kim, and Chang~D Yoo.
  2019{\natexlab{a}}.
\newblock Gaining extra supervision via multi-task learning for multi-modal
  video question answering.
\newblock In \emph{2019 International Joint Conference on Neural Networks
  (IJCNN)}, pages 1--8. IEEE.

\bibitem[{Kim et~al.(2019{\natexlab{b}})Kim, Ma, Kim, Kim, and
  Yoo}]{kim2019progressive}
Junyeong Kim, Minuk Ma, Kyungsu Kim, Sungjin Kim, and Chang~D Yoo.
  2019{\natexlab{b}}.
\newblock Progressive attention memory network for movie story question
  answering.
\newblock In \emph{Proceedings of the IEEE Conference on Computer Vision and
  Pattern Recognition}, pages 8337--8346.

\bibitem[{Kim et~al.(2017)Kim, Heo, Choi, and Zhang}]{kim2017deepstory}
Kyung-Min Kim, Min-Oh Heo, Seong-Ho Choi, and Byoung-Tak Zhang. 2017.
\newblock Deepstory: video story qa by deep embedded memory networks.
\newblock In \emph{Proceedings of the 26th International Joint Conference on
  Artificial Intelligence}, pages 2016--2022. AAAI Press.

\bibitem[{Kingma and Ba(2015)}]{kingma2014adam}
Diederik~P Kingma and Jimmy Ba. 2015.
\newblock Adam: A method for stochastic optimization.
\newblock In \emph{3rd International Conference on Learning Representations,
  {ICLR} 2015,San Diego, CA, USA, May 7-9, 2015, Conference Track Proceedings}.

\bibitem[{Krause et~al.(2017)Krause, Johnson, Krishna, and
  Fei-Fei}]{krause2017hierarchical}
Jonathan Krause, Justin Johnson, Ranjay Krishna, and Li~Fei-Fei. 2017.
\newblock A hierarchical approach for generating descriptive image paragraphs.
\newblock In \emph{Computer Vision and Pattern Recognition (CVPR), 2017 IEEE
  Conference on}, pages 3337--3345. IEEE.

\bibitem[{Lei et~al.(2018)Lei, Yu, Bansal, and Berg}]{lei2018tvqa}
Jie Lei, Licheng Yu, Mohit Bansal, and Tamara~L Berg. 2018.
\newblock Tvqa: Localized, compositional video question answering.
\newblock In \emph{EMNLP}.

\bibitem[{Lei et~al.(2020)Lei, Yu, Berg, and Bansal}]{lei2019tvqa+}
Jie Lei, Licheng Yu, Tamara~L Berg, and Mohit Bansal. 2020.
\newblock Tvqa+: Spatio-temporal grounding for video question answering.
\newblock \emph{Proceedings of the 58th Annual Meeting of the Association for
  Computational Linguistics}.

\bibitem[{Li et~al.(2019)Li, Wang, Shen, and Hengel}]{Li_2019_CVPR}
Hui Li, Peng Wang, Chunhua Shen, and Anton van~den Hengel. 2019.
\newblock Visual question answering as reading comprehension.
\newblock In \emph{The IEEE Conference on Computer Vision and Pattern
  Recognition (CVPR)}.

\bibitem[{Liang et~al.(2017)Liang, Hu, Zhang, Gan, and
  Xing}]{liang2017recurrent}
Xiaodan Liang, Zhiting Hu, Hao Zhang, Chuang Gan, and Eric~P Xing. 2017.
\newblock Recurrent topic-transition gan for visual paragraph generation.
\newblock In \emph{Proceedings of the IEEE International Conference on Computer
  Vision}, pages 3362--3371.

\bibitem[{Liu et~al.(2019)Liu, Ott, Goyal, Du, Joshi, Chen, Levy, Lewis,
  Zettlemoyer, and Stoyanov}]{liu2019roberta}
Yinhan Liu, Myle Ott, Naman Goyal, Jingfei Du, Mandar Joshi, Danqi Chen, Omer
  Levy, Mike Lewis, Luke Zettlemoyer, and Veselin Stoyanov. 2019.
\newblock Roberta: A robustly optimized bert pretraining approach.
\newblock \emph{arXiv preprint arXiv:1907.11692}.

\bibitem[{Lu et~al.(2016)Lu, Yang, Batra, and Parikh}]{lu2016hierarchical}
Jiasen Lu, Jianwei Yang, Dhruv Batra, and Devi Parikh. 2016.
\newblock Hierarchical question-image co-attention for visual question
  answering.
\newblock In \emph{Advances In Neural Information Processing Systems}, pages
  289--297.

\bibitem[{Maharaj et~al.(2017)Maharaj, Ballas, Rohrbach, Courville, and
  Pal}]{maharaj2017dataset}
Tegan Maharaj, Nicolas Ballas, Anna Rohrbach, Aaron Courville, and Christopher
  Pal. 2017.
\newblock A dataset and exploration of models for understanding video data
  through fill-in-the-blank question-answering.
\newblock In \emph{Proceedings of the IEEE Conference on Computer Vision and
  Pattern Recognition}, pages 6884--6893.

\bibitem[{Melas-Kyriazi et~al.(2018)Melas-Kyriazi, Rush, and
  Han}]{melaskyriazi2018paragraph}
Luke Melas-Kyriazi, Alexander Rush, and George Han. 2018.
\newblock Training for diversity in image paragraph captioning.
\newblock \emph{EMNLP}.

\bibitem[{Pennington et~al.(2014)Pennington, Socher, and
  Manning}]{pennington2014glove}
Jeffrey Pennington, Richard Socher, and Christopher Manning. 2014.
\newblock Glove: Global vectors for word representation.
\newblock In \emph{Proceedings of the 2014 conference on empirical methods in
  natural language processing (EMNLP)}, pages 1532--1543.

\bibitem[{Seo et~al.(2017)Seo, Kembhavi, Farhadi, and
  Hajishirzi}]{seo2016bidirectional}
Min~Joon Seo, Aniruddha Kembhavi, Ali Farhadi, and Hannaneh Hajishirzi. 2017.
\newblock Bidirectional attention flow for machine comprehension.
\newblock In \emph{ICLR}.

\bibitem[{Shou et~al.(2017)Shou, Chan, Zareian, Miyazawa, and
  Chang}]{shou2017cdc}
Zheng Shou, Jonathan Chan, Alireza Zareian, Kazuyuki Miyazawa, and Shih-Fu
  Chang. 2017.
\newblock Cdc: Convolutional-de-convolutional networks for precise temporal
  action localization in untrimmed videos.
\newblock In \emph{Proceedings of the IEEE Conference on Computer Vision and
  Pattern Recognition}, pages 5734--5743.

\bibitem[{Shou et~al.(2016)Shou, Wang, and Chang}]{shou2016temporal}
Zheng Shou, Dongang Wang, and Shih-Fu Chang. 2016.
\newblock Temporal action localization in untrimmed videos via multi-stage
  cnns.
\newblock In \emph{Proceedings of the IEEE Conference on Computer Vision and
  Pattern Recognition}, pages 1049--1058.

\bibitem[{Tapaswi et~al.(2016)Tapaswi, Zhu, Stiefelhagen, Torralba, Urtasun,
  and Fidler}]{tapaswi2016movieqa}
Makarand Tapaswi, Yukun Zhu, Rainer Stiefelhagen, Antonio Torralba, Raquel
  Urtasun, and Sanja Fidler. 2016.
\newblock Movieqa: Understanding stories in movies through question-answering.
\newblock In \emph{Proceedings of the IEEE conference on computer vision and
  pattern recognition}, pages 4631--4640.

\bibitem[{Weinzaepfel et~al.(2015)Weinzaepfel, Harchaoui, and
  Schmid}]{weinzaepfel2015learning}
Philippe Weinzaepfel, Zaid Harchaoui, and Cordelia Schmid. 2015.
\newblock Learning to track for spatio-temporal action localization.
\newblock In \emph{Proceedings of the IEEE international conference on computer
  vision}, pages 3164--3172.

\bibitem[{Wu et~al.(2019)Wu, Hu, and Mooney}]{wu2019generating}
Jialin Wu, Zeyuan Hu, and Raymond Mooney. 2019.
\newblock Generating question relevant captions to aid visual question
  answering.
\newblock In \emph{Proceedings of the 57th Annual Meeting of the Association
  for Computational Linguistics}, pages 3585--3594.

\bibitem[{Xu and Saenko(2016)}]{xu2016ask}
Huijuan Xu and Kate Saenko. 2016.
\newblock Ask, attend and answer: Exploring question-guided spatial attention
  for visual question answering.
\newblock In \emph{European Conference on Computer Vision}, pages 451--466.
  Springer.

\bibitem[{Yang et~al.(2017)Yang, Tang, Yang, and Li}]{yang2017dense}
Linjie Yang, Kevin Tang, Jianchao Yang, and Li-Jia Li. 2017.
\newblock Dense captioning with joint inference and visual context.
\newblock In \emph{Proceedings of the IEEE Conference on Computer Vision and
  Pattern Recognition}, pages 2193--2202.

\bibitem[{Yang et~al.(2016)Yang, He, Gao, Deng, and Smola}]{yang2016stacked}
Zichao Yang, Xiaodong He, Jianfeng Gao, Li~Deng, and Alex Smola. 2016.
\newblock Stacked attention networks for image question answering.
\newblock In \emph{Proceedings of the IEEE Conference on Computer Vision and
  Pattern Recognition}, pages 21--29.

\bibitem[{Zadeh et~al.(2019)Zadeh, Chan, Liang, Tong, and
  Morency}]{zadeh2019social}
Amir Zadeh, Michael Chan, Paul~Pu Liang, Edmund Tong, and Louis-Philippe
  Morency. 2019.
\newblock Social-iq: A question answering benchmark for artificial social
  intelligence.
\newblock In \emph{Proceedings of the IEEE Conference on Computer Vision and
  Pattern Recognition}, pages 8807--8817.

\bibitem[{Zhu et~al.(2016)Zhu, Groth, Bernstein, and Fei-Fei}]{zhu2016visual7w}
Yuke Zhu, Oliver Groth, Michael Bernstein, and Li~Fei-Fei. 2016.
\newblock Visual7w: Grounded question answering in images.
\newblock In \emph{Proceedings of the IEEE Conference on Computer Vision and
  Pattern Recognition}, pages 4995--5004.

\end{thebibliography}
\bibliographystyle{acl_natbib}

\end{document}